\documentclass[10pt,journal,compsoc,nocompress]{IEEEtran}
\usepackage{amsmath}
\usepackage{array}
\usepackage{longtable}
\usepackage{filecontents,lipsum}
\usepackage{blindtext}
\usepackage[utf8]{inputenc}
\usepackage[english]{babel}
\usepackage{filecontents,lipsum}
\usepackage[dvipsnames]{xcolor}
\usepackage{amssymb}
\usepackage{times}
\usepackage[noend]{algpseudocode}
\usepackage[ruled]{algorithm}
\usepackage{algpseudocode}
\usepackage{algorithm}
\usepackage{caption}
\usepackage{lineno}
\usepackage{ntheorem}
\usepackage{graphicx}
\usepackage{tikz}
\usepackage{multirow}
\usepackage{pgfplots}
\usetikzlibrary{
                chains,
                positioning,
                shapes.geometric,
                shapes,arrows,
                patterns,
                decorations.markings
                }
\usepackage{subcaption}
\usepackage{tikz}
\usepackage{hyperref}
\usepackage{multicol}

\pgfplotsset{compat=1.14}
\begin{document}
\title{Accurate and fast matrix factorization for low-rank learning}

\author{Reza~Godaz,
        Reza~Monsefi,
        Faezeh~Toutounian,
        Reshad~Hosseini% <-this % stops a space
\IEEEcompsocitemizethanks{\IEEEcompsocthanksitem R. Godaz and R. Monsefi are with the Department of Computer Engineering, Ferdowsi University of Mashhad (FUM), Mashhad, Iran.\protect\\
E-mail: reza.godaz@mail.um.ac.ir ; monsefi@um.ac.ir
\IEEEcompsocthanksitem F. Toutounian is with the Department of Applied Mathematics, School of Mathematical Sciences, Ferdowsi University of Mashhad, Mashhad, Iran.\protect\\
E-mail:  toutouni@math.um.ac.ir.
\IEEEcompsocthanksitem R. Hosseini is with the Department of Electrical and Computer Engineering, University of Tehran, Teharn, Iran.\protect\\
E-mail: reshad.hosseini@ut.ac.ir.}% <-this % stops an unwanted space
%\thanks{Manuscript received March - , 2021; }
}

\IEEEtitleabstractindextext{
\begin{abstract}
In this paper, we tackle two important problems in low-rank learning, which are partial singular value decomposition  and numerical rank estimation of huge matrices. By using the concepts of Krylov subspaces such as  Golub-Kahan bidiagonalization (GK-bidiagonalization) as well as Ritz vectors, we propose two methods for solving these problems in a fast and accurate way. Our experiments show the advantages of the proposed methods compared to the traditional and randomized singular value decomposition methods.
The proposed methods are appropriate for applications involving huge matrices where the accuracy of the desired singular values and also all of their corresponding singular vectors are essential. As a real application, we evaluate the performance of our methods on the problem of Riemannian similarity learning between two various image datasets of MNIST and USPS.
\end{abstract}

}

\maketitle

\IEEEdisplaynontitleabstractindextext

% \fi
\IEEEpeerreviewmaketitle

\IEEEraisesectionheading{\section{Introduction}\label{sec:introduction}}

%%%%%%%%%%%%%  Why low-rank?   %%%%%%%%%%%%%%%%
\IEEEPARstart{S}{ingular}
Singular value decomposition (SVD) \cite{golub1965calculating,golub1971singular} is an important matrix factorization method with a wide variety of applications in various sciences such as mathematics, artificial intelligence and physics. Common methods for solving SVD can be used when the input matrix is small or at-most medium sized, but they can not be used (or require excessive execution time) for problems that involve huge matrices (matrices with more than $1e8$ entries). There are some methods that use randomization and sampling to tackle this problem \cite{martinsson2011randomized,halko2011finding,martinsson2019randomized}.
For example, fast SVD algorithms of \cite{halko2011finding} are based on randomization and sampling, and have strong theoretical gaurantees. These methods are currently used in various fields like social network data analysis \cite{Facebook:2014:Online}.
In sampling and randomization methods, the large input matrix is converted into a smaller matrix considering a predefined error, and then one of the standard SVD methods can be applied. As a result, due to the use of the standard SVD, these methods are obliged to find a trade-off between determining the column sampling rate (that is related to the accuracy), and the execution time. Our experiments show that the eigenvalues and corresponding eigenvectors obtained by these algorithms are not accurate, although the predefined errors are maintained (see Section \ref{section_exp_results}).

A fundamental problem in machine learning related to SVD is modeling vast amounts of data in a low-dimensional representation. Processing raw data in typical ways is time-consuming and not always feasible, especially when the data are high dimensional. Representing data in a suitable lower-dimensional form can alleviate challenges of computational complexity, high memory usage, model compression, and noisy environments\cite{cambier2016robust}. The reduced dimensional form of the data can be used in a wide variety of applications such as neural language processing \cite{dhillon2015eigenwords}, video and image processing \cite{zhou2014low}, recommender systems \cite{vandereycken2013low,cambier2016robust}, multidimensional scaling \cite{bartell1992latent,qiu2020unsupervised}, collaborative filtering and genomics data in bioinformatics \cite{bennett2007netflix,rennie2005fast}. It has been shown that in huge matrices the numerical rank of the matrix grows logarithmically with its dimensions \cite{udell2019big}.
The rank on the data matrix has different interpretations in different applications. For example in latent semantic analysis, the rank can be interpreted as the number of concepts related to a document. In recommender systems, the rank of a user-rating matrix corresponds to a small set of typical users \cite{Markovsky2019}.

For problems involving factorization of huge matrices, there is currently a significant gap between being fast, and having high level of accuracy. In the current work, the aim is to bridge this gap by using the well-known Krylov subspace methods and Ritz vectors  \cite{golub1965calculating,sorensen2002numerical}.
In what follows, our aim is to pursue the following two goals: (i) to find a fixed number of singular values and corresponding singular vectors of a huge matrix in order to transform the matrix into a low-rank form accurately with moderate time and memory usage, (ii) to approximate the rank of a huge matrix in a fast and accurate way.

The remainder of this article is organized as follows. Section \ref{sec_background} contains a brief review of the SVD problem, and basic concepts such as Ritz vectors and  Golub-Kahan bidiagonalization (GK-bidiagonalization). Our algorithm for accurate and fast SVD is the focus of Section \ref{section_fast_svd}. In this section, an algorithm to determine an accurate numerical rank of a matrix is also presented. Section \ref{section_app_data_pair} introduces an application of the proposed method in Riemannian similarity learning (RSL). In Section \ref{section_experiments}, we evaluate he efficiency of the proposed algorithms on several numerical examples. Finally, we make some concluding remarks in Section \ref{section_conclusion}.

\section{Background}
\label{sec_background}
In this section, we give an overview of SVD and randomized methods to solve SVD. Then, we introduce Ritz vectors and GK-bidiagonalization. In the next section, we should that truncated GK-bidiagonalization can be used for approximate computation of the largest singular values and the rank of a matrix.
\subsection{Singular value decomposition}\label{sec_svd}
SVD is a well-known mathematical method that has a wide variety of real world applications in machine learning and mathematical problems.
SVD of matrix $A$ with size $m\times n$ and rank $r$ is given by\cite{golub1965calculating,golub1971singular}:
\begin{equation}
A=U{\Sigma}V^T, \qquad \sum=\left [\begin{array}{ll} D & 0\\0 & 0\end{array}\right ],
\label{eq_svd}
\end{equation}
where $U\in \mathbb{R}^{m \times m}$ and $V\in \mathbb{R}^{n \times n}$ are matrices with orthogonal columns, and ${D}={\rm diag}(\sigma_1,  \sigma_2,\ldots, \sigma_r)$ is a diagonal matrix. Here, $\sigma_1\geq \sigma_2\geq ...\geq\sigma_r>0$ denote  $r\leq \textbf{min}(m, n)$ nonzero singular values of $A$. The computational complexity of traditional SVD methods for matrix $A\in \mathbb{R}^{m\times n}$ is $\mathcal{O}(mn\textbf{ min}(m,n))$. An important use of this algorithm in learning problems is related to finding a low-rank form of the data matrix or a parameter matrix, where we are looking for  $k$ singular vectors associated with the $k$ largest singular values of the matrix.

\subsubsection{Randomization-based methods}
\label{section_randomized}
Randomized methods \cite{halko2011finding,martinsson2019randomized} can make it possible to compute singular values and corresponding singular vectors of huge matrices in a reasonable amount of time. These methods usually use Eckart-Young theorem \cite{stewart1993early} to find the rank of the input matrix, and thereby determine the required number of sampled columns (see Section 4 in \cite{martinsson2019randomized} and also Section 10.2 in \cite{halko2011finding}). In \cite{halko2011finding}, the authors added this a fixed number (called oversampling parameter) to this theoretically-driven number to get the final sampling rate.

Experiments of Section \ref{section_experiments} demonstrate that when the input matrix is huge, setting a small value for oversampling parameter (for example 10 as suggested by authors in \cite{halko2011finding}) results in a fast execution time but inaccurate results; whereas fixing a larger value for it implies finding more accurate results, but the execution time will inevitably be longer. Based on the results given in \cite{halko2011finding} (and also our experiments), certain factors affect this oversampling parameter, such as the dimension, the singular values, and the numerical rank of the input matrix. Since both the numerical rank and the singular values of the matrix are unknown before setting the oversampling parameter (since they are costly to be computed), the resulting decomposition is not very accurate in huge matrices having a large numerical rank.

\subsection{Golub-Kahan bidiagonalization for computing Ritz Vectors }
\label{sec_gk_bidiag}
We first recall the definition of a  Ritz pair \cite{sorensen2002numerical}.
Then, we describe the GK-bidiagonalization process for computing  the $l$ largest Ritz values and the corresponding Ritz vectors of $A^TA$ and $AA^T$ \cite{baglama2013augmented,golub1965calculating,sorensen2002numerical}.

%\noindent
{\bf Definition 1.}
Consider a k-dimensional subspace $\mathcal{W}$ of $\mathbb{C}^n$. Given a
matrix $B \in \mathbb{C} ^{n\times n}, \tilde{\theta} \in \mathbb{C}$, and $\tilde{y} \in \mathcal{W}$, $(\tilde{\theta}, \tilde{y})$ is a Ritz pair of $B$ with respect to $\mathcal{W}$ if and only if
$$B\tilde{y} - \tilde{\theta} \tilde{y} \perp \mathcal{W} $$
or equivalently, for the canonical scalar product,
$$\forall w \in \mathcal{W},~~~~ w^H (B\tilde{y} - \tilde{\theta}\tilde{y}) = 0.$$
We call $\tilde{y}$ a  Ritz vector associated with the  Ritz value $\tilde{\theta}$.

 Let $W_k=[w_1,w_2,\ldots,w_k]\in \mathbb{R}^{n\times k} $ be an orthogonal basis  of subspace   $ \mathcal{W} $. Using Definition 1,  the  Ritz pairs $ (\tilde{\theta}_i,\tilde{y}_i), i=1,2,\ldots,k, $ of $A^TA$ and $AA^T$ can be obtained by solving  the following small eigenvalue problems
\begin{align}\label{eq5}
W_k^TA^TAW_kg_i=\tilde{\theta}_ig_i,\quad \tilde{y}_i=W_kg_i,\quad i=1,2,\ldots,k
\end{align}
and
\begin{align}\label{eq6}
W_k^TAA^TW_kg_i=\tilde{\theta}_ig_i,\quad \tilde{y}_i=W_kg_i,\quad i=1,2,\ldots,k,
\end{align}
respectively. As it is known \cite[Pages~215-217]{berger1981symmetric}, this procedure is optimal in a global sense and is optimal for exterior eigenvalues.

We begin by recalling some necessary properties of the  GK-bidiagonalization process \cite{golub1965calculating} which is one of the well known
algorithms for computing the orthonormal bases for Krylov subspaces
\begin{equation} \label{krylov}
\begin{aligned}
\mathcal{K}_m(AA^T,q_1)=span\{q_1,(AA^T)^1q_1,...,(AA^T)^{k}q_1\},\\
\mathcal{K}_m(A^TA,p_1)=span\{p_1,(A^TA)^1p_1,...,(A^TA)^{k}p_1\},
\end{aligned}
\end{equation}
with $q_1=q_1/\|q_1\|$ and $p_1=A^Tq_1/\|A^Tq_1\|$, where $q_1\in \mathbb{R}^m$ is a  nonzero arbitrary vector and  $\|.\|$ denotes the Euclidean vector norm or associated matrix norm.

Assuming that $A$ has full column rank and $k\ll\textbf{min}(m,n)$, the GK-bidiagonalization process \cite{golub1965calculating} which is called the procedure Bidiag 1 in \cite{paige1982lsqr}, reduces $A$ to its lower bidiagonal form. Procedure Bidiag 1 with starting vector $q_{1}$ can be described as follows:

\begin{equation}\label{eq2.1}
 \hspace{-2cm} \beta_{1}q_1 = q_1, ~~~ \alpha_{1}p_1 = A^Tq_1,
\end{equation}
\begin{equation}\label{eq2.2}
\begin{array}{ll}
\left .
\begin{array}{l}
\beta_{i+1}q_{i+1} = Ap_i- \alpha_{i}q_i,\\
\alpha_{i+1}p_{i+1} = A^Tq_{i+1}- \beta_{i+1}p_i,
\end{array} \right \} & i = 1, 2, \ldots  .
\end{array}
\end{equation}
The scalars $\alpha_{i}\geq 0$ and $\beta_{i}\geq 0$ are chosen such that $\|q_{i}\|_2 = \|p_{i}\|_2 =1$. Together with the definitions
\begin{equation*}
Q_k=  [q_1, q_2,\ldots, q_k], ~~~~~ P_k=[p_1, p_2,\ldots, p_k],
\end{equation*}
and
\begin{equation}\label{eq2.3}
 {B}_{k+1,k}=
\left [
\begin{array}{llll}
\alpha_{1} & & &\\
\beta_2 & \alpha_2 & & \\
 &\ddots &\ddots &\\
 & & \beta_k & \alpha_k\\
 & & & \beta_{k+1}
 \end{array} \right ],
 \end{equation}

\noindent the recurrence relations \eqref{eq2.1} and \eqref{eq2.2} can be rewritten as the following equations:
\begin{equation}\label{eq2.4}
 \begin{array}{rl}
Q_{k+1}(\beta_1e_1) =&q_{1},\\
AP_k =&Q_{k+1}B_{k+1,k},\\
A^TQ_{k+1} =&P_kB^T_{k+1,k} + \alpha_{k+1}p_{k+1}e^T_{k+1},
\end{array}
\end{equation}
where $e_1$ and $e_{k+1}$ denote the first and the $(k+1)$st columns of an identity
matrix of order $k+1$, respectively. In exact arithmetic, we have
\begin{equation}\label{eq_pq_ortho}
 \begin{array}{rl}
Q_{k+1}^{T}Q_{k+1}=&I,\\
P_{k}^{T}P_{k}=&I,
\end{array}
\end{equation}
where $I$ is the identity matrix. The columns of $Q_{k}$ and $P_{k}$ are orthonormal bases for the Krylov subspaces defined in \eqref{krylov}.

\begin{algorithm}[t]
\caption{GK-bidiagonalization and numerical rank estimation} \label{alg_fast_rank}

\textbf{Inputs:}
\\\hspace*{\algorithmicindent}$A \in \mathbb{R}^{m \times n}$,
%\\\hspace*{\algorithmicindent}$q_1 \in \mathbb{R}^m$,
\\\hspace*{\algorithmicindent}$k$: number of iterations ($k\leq \textbf{min}(m,n)$),
\\\hspace*{\algorithmicindent}$\epsilon \in \mathbb{R}$.
\\\textbf{Outputs:}
\\\hspace*{\algorithmicindent} $k'=\textbf{min}(k,\text{ the approximate numerical rank of A})$,
\\\hspace*{\algorithmicindent}$B_{k'+1,k'} \in \mathbb{R}^{(k'+1) \times (k')}$,
\\\hspace*{\algorithmicindent}$P_{k'} \in \mathbb{R}^{n \times k'}$,
%\\\hspace*{\algorithmicindent}$p_{(j+1)} \in \mathbb{R}^m$
\\\hspace*{\algorithmicindent}$Q_{k'+1} \in \mathbb{R}^{m \times (k'+1)}$,
%\\\hspace*{\algorithmicindent}$\alpha_{k'+1} \in \mathbb{R}$.
\\\hspace*{\algorithmicindent}

\begin{algorithmic}[1]
\State $q_1\sim\mathcal{N}(2,1)^{m\times 1}$ ,  $\beta_1=\|q_1\|$ , $q_1=q_1/\beta_1$,$Q_1=q_1$
\State $p_1=A^Tq_1$ , $\alpha_1=\|p_1\|$ , $p_1=p_1/\alpha_1$ , $P_1=p_1$
\State $k'=0$
\State  while $k'<k$ do
\State \hspace*{\algorithmicindent}$q_{k'+1}=Ap_{k'}-q_{k'}\alpha_{k'}$
\State \hspace*{\algorithmicindent}$q_{k'+1}=q_{k'+1}-Q_{(1:k')}(Q_{(1:k')}^Tq_{k'+1})$
\State \hspace*{\algorithmicindent}$\beta_{k'+1}=\|q_{k'+1}\|$
\State \hspace*{\algorithmicindent}$q_{k'+1}=q_{k'+1}/\beta_{k'+1}$ , $Q_{k'+1}=[Q_{k'}$ , $q_{k'+1}]$
\State \hspace*{\algorithmicindent} if $\|q_{k'+1}\|<\epsilon$ do
%\State \\

%\hspace*{\algorithmicindent}\hspace*{\algorithmicindent}$r=j$
%\State \\ \hspace*{\algorithmicindent}\hspace*{\algorithmicindent}Arrange $B$ matrix as $B[1:r+1,1:r]$
%\State \\ \hspace*{\algorithmicindent}\hspace*{\algorithmicindent}Arrange $Q$ matrix as $Q[:,1:rank+1]$
%\State \\ \hspace*{\algorithmicindent}\hspace*{\algorithmicindent}Arrange $P$ matrix as $P[:,1:rank]$
\State \hspace*{\algorithmicindent}\hspace*{\algorithmicindent}break
\State \hspace*{\algorithmicindent}end if
\State \hspace*{\algorithmicindent}$ p_{k'+1}= A^T q_{k'+1}-p_j\beta_{k'+1}$
\State \hspace*{\algorithmicindent}$p_{k'+1}=p_{k'+1}-P_{(1:k')}(P_{(1:k')}^Tp_{k'+1})$
\State \hspace*{\algorithmicindent}$\alpha_{k'+1}=\|p_{k'+1}\|$ , $p_{k'+1}=p_{k'+1}/\alpha_{k'+1}$
\State \hspace*{\algorithmicindent}$P_{k'+1}=[P_{k'},p_{k'+1}]$
\State \hspace*{\algorithmicindent}$k'=k'+1$
\State end while
\end{algorithmic}
\end{algorithm}

\section{Accurate and fast SVD (F-SVD) algorithm}
\label{section_fast_svd}
The largest Ritz values and the corresponding Ritz vectors of $A^TA$ and $AA^T$ can be used for computing  $l$ singular triplets $\{ {\sigma}_i,u_i,v_i \}_{i=1}^l$ associated with the approximate $l$ largest singular values of $A$. Therefore, GK-bidiagonalization can be used to find the approximation of singular values and corresponding singular vectors of an input matrix. The  GK-bidiagonalization process is outlined in Algorithm~\ref{alg_fast_rank}, that can also be used to estimate the numerical rank of a matrix. The vectors $q_{k'+1}$ and $p_{k'+1}$ in lines 6 and 13 of the algorithm are re-orthogonalized to preserve orthogonality. Additional details containing the GK-bidiagonalization process can be found in \cite{paige1982lsqr}. Multiplying the second  equation in (\ref{eq2.4}) by $A^T$ yields
\begin{equation}\label{eq2.10}
A^TAP_k= P_kB^T_{k+1,k}B_{k+1,k} + \alpha_{k+1}\beta_{k+1}p_{k+1}e^T_k.
\end{equation}
We wish to find approximate eigenvectors of $A^TA$ from the subspace spanned by the columns of $P_k$.
To compute the approximate eigenvectors of $A^TA$, by using relation (\ref{eq2.10}), the small  eigenvalue problem posed at (\ref{eq5}) can be written as follows
\begin{equation}\label{eq2.13}
B_{k+1,k}^TB_{k+1,k}g_i=\tilde{\theta}_ig_i,\quad \tilde{v}_i=P_kg_i,\quad i=1,2,\ldots,k.
\end{equation}
By using \eqref{eq_svd}, the approximate right singular vector $\tilde{u}_i,$ $i={1,2,...,k,}$ can be computed as follows:
\begin{equation}\label{svdV}
\tilde{u}_i=\frac{1}{\sigma_i}A\tilde{v}_i^T,\quad i=1,2,...,k.
\end{equation}
The resulting algorithm is summarized as Algorithm~\ref{alg_fast_SVD}.

\begin{algorithm}[thbp]
\caption{Accurate and fast SVD (F-SVD)}\label{alg_fast_SVD}

\textbf{Inputs:}
\\\hspace*{\algorithmicindent}$A \in \mathbb{R}^{m \times n}$,
\\\hspace*{\algorithmicindent}$q$ is arbitrary,
\\\hspace*{\algorithmicindent}$k$: number of the iterations in Algorithm \ref{alg_fast_rank},
\\\hspace*{\algorithmicindent}$r$: number of desired larger eigen triplets $\{\sigma_i,u_i,v_i\}$.
\\\textbf{Output:}
\\\hspace*{\algorithmicindent}$r$ desired larger eigen triplets $\{\sigma_i,u_i,v_i\}$ of A.
\\\hspace*{\algorithmicindent}

\begin{algorithmic}[1]
\State Set $k$ as number of iterations and run Algorithm~\ref{alg_fast_rank} to find $B_{k'+1,k'}$,$P_{k'}$,$Q_{k'+1}$
\State Find eigen decomposition of $(B_{k'+1,k'}^TB_{k'+1,k'})$ for $V_1$ (eigenvectors) and $S_1$ (eigenvalues)
\State $V_2=P_{k'}V_1$
\State Select $r$ larger eigenvalues of $S_1$ and corresponding eigenvectors from $V_2$ as $\Sigma_1$ and $V_r$, respectively
\State $\Sigma_r=\sqrt{\Sigma_1}$
\State for $i=1$ to $r$ do
\State \hspace*{\algorithmicindent} $U_r[:,i]=1/\sigma_i AV_r[:,i]$
\State end for
\State Return $U_r$, $\Sigma_r$ and $V_r$.
\end{algorithmic}
\end{algorithm}

\subsection{Complexity analysis  and comparison with randomized SVD \cite{halko2011finding}}
\label{section_complexity_F-SVD}
The maximal computational complexity of the first three lines of Algorithm~\ref{alg_fast_rank} is related to computing $A^Tq_1$ that is $\mathcal{O}(mn)$ flops. Computing $Ap_{k'}$ in line 5, and also $A^Tq_{k'+1}$ in line 12, both takes $\mathcal{O}(mnk')$ flops (concerning the loop), and it is $\mathcal{O}(m{k'}^2)$ and $\mathcal{O}(n{k'}^2)$ flops for lines 6 and 13 respectively. Therefore the overall computational complexity involves $\mathcal{O}(mnk'+(m+n)k'^2)$ flops.

With respect to the memory requirements of Algorithm \ref{alg_fast_rank}, in addition to the input matrix that can be sparse or dense, we consider the $P$ and $Q$ matrices inside the algorithm each require memory of order $\mathcal{O}((m+n)k')$. Because matrix $B$ is a bidiagonal matrix, two vectors of length $k'$ can be used for it. As a result the overall memory usage of Algorithm 1 is $\mathcal{O} ((m+n+2)k')$ which is gain of order $\mathcal{O}((m+n)k')$.

The first part of Algorithm \ref{alg_fast_SVD} involves computing $B_{k'+1,k'}$, $P_{k'}$, and $Q_{k'+1}$ (using Algorithm \ref{alg_fast_rank}) which has computational complexity $\mathcal{O}(mnk'+(m+n)k'^2)$ flops. Then, the eigen decomposition of the matrix $B_{k'+1,k'}^TB_{k'+1,k'}$ in the second step is computed. The computational complexity of this computation, in general requires at most $\mathcal{O}({k'}^3)$ flops. In our algorithms, because the matrix $B_{k'+1,k'}^TB_{k'+1,k'}$ is a small tridiagonal matrix, the computational complexity of this part of the algorithm is close to $\mathcal{O}(k'^2)$. We also have a matrix multiplication in step 3 that involves $\mathcal{O}(m{k'}^2)$ flops. In steps 6 and 7, there is a loop which requires $\mathcal{O}(mnr)$ flops. As a result, the overall complexity of this algorithm amounts to $\mathcal{O}(mn(k'+r)+(m+n)k'^2)$ flops. Using the assumption $k',r\ll min(m,n)$ (that is correct when the numerical rank of the input matrix is small) the computational complexity of Algorithm \ref{alg_fast_SVD} is $\mathcal{O}(mnk')$.

Suppose that $X\in \mathbb{R}^{m\times n}$ and we are looking for a low-rank form for $X$ with the numerical rank $k$. Using relation \eqref{eq_svd}, the computational complexity of the traditional SVD is $\mathcal{O}(mn^2),n<m$ and $\mathcal{O}(mn\log(l)+l^2(m+n)),\quad l=k+p$ for the randomized SVD (R-SVD) \cite{halko2011finding}. In \cite{halko2011finding}, the authors assumed that the oversampling parameter $p$ is a small value that can be ignored. But for the matrices where the decay of the singular values is slow, the oversampling parameter is not a small value and cannot be ignored. In such situations, the second term of the computational complexity of the randomized algorithm becomes dominant. Therefore its computational complexity approaches that of the traditional method of SVD.

As we observe, the standard SVD, R-SVD, and F-SVD methods can be used for determining the k
largest singular values of a matrix, but in different qualities. The standard SVD computes a complete singular decomposition and the results are accurate, but it is not reasonable when $k\ll \min(m,n)$ because of memory and time requirements. The R-SVD method, as explained before, selects a subset of  columns of $A$ and performs a complete SVD on this subset. So its time and memory requirements reduce dramatically, but the resulted singular values and corresponding singular vectors are not accurate. For this purpose, by selecting a suitable value for k, the F-SVD method is able to provide the  accurate results.  In addition, its execution time and memory usage are comparable to those of other methods.  The numerical experiments (Section \ref{section_experiments}) demonstrate the efficiency of  the F-SVD Algorithm and confirm that this algorithm  quickly executes and provides the accurate results for huge matrices.

\subsection{Fast numerical rank estimation}
\label{section_fast_rank_estimation}

To estimate the numerical rank of an input matrix $A\in \mathbb{R}^{m\times n}$, we use the theory of the Lanczos process in applying the criterion $\|q_{k'+1}\|< \epsilon$ to terminate the Algorithm \ref{alg_fast_rank} \cite{lanczos1950iteration,paige1982lsqr}. In this formulation $\epsilon$ is a very small positive value determined by the user. This criterion prevents the algorithm from performing extra iterations. As a result, if this condition is satisfied after $k'$ iterations and $k'<k$, then $k'$ can be used as a preliminary estimate of the numerical rank of the matrix, but this estimate is not very accurate (see Algorithm~\ref{alg_fast_rank}). Thereafter, one can obtain an accurate numerical rank using eigen decomposition of the matrix $B_{k'+1,k'}^TB_{k'+1,k'}$ generated in Algorithm \ref{alg_fast_rank}; see Algorithm \ref{alg_rank_determination} for details.

\begin{algorithm}
\caption{Rank determination algorithm}
\label{alg_rank_determination}

\textbf{Inputs:}
\\\hspace*{\algorithmicindent}$A \in \mathbb{R}^{m \times n}$,
\\\hspace*{\algorithmicindent}$\epsilon$: a small value (default: 1e-8).
\\\textbf{Output:}
\\\hspace*{\algorithmicindent} $r$: The rank of A.
\\\hspace*{\algorithmicindent}

\begin{algorithmic}[1]
\State $k=min(m,n)$
\State Set $k$ as the number of iterations and run Algorithm~\ref{alg_fast_rank} to find $B_{k'+1,k'}$
\State Find the eigen decomposition of $(B_{k'+1,k'}^TB_{k'+1,k'})$ to determine $S$, the matrix that contains the eigenvalues.
\State Count the diagonal elements of $S$ that exceed $\epsilon$ as the accurate numerical rank $r$.
\end{algorithmic}
\end{algorithm}

\section{An application: Riemannian similarity learning}
\label{section_app_data_pair}
There are a wide variety of learning problems in which the parameters of interest lie on a Riemannian manifold, such as principal components analysis (PCA) \cite{wold1987principal} and robust PCA \cite{candes2011robust,zhang2018robust}, independent component analysis (ICA) \cite{hyvarinen2000independent}, subspace learning, low-rank matrix completion  \cite{vandereycken2013low}, dictionary learning \cite{Cherian2015RiemannianDL} and Gaussian mixture model \cite{hosseini2015matrix}.
One of the important assumptions in some of these learning problems is that the data are localized around low-dimensional structures, that is we have Riemannian manifold of low-rank matrices. When the optimization constraint is having low-rank matrices, we can use Riemannian optimization algorithm \cite{vandereycken2013low}. In-depth discussions of different Riemannian optimization methods with their theoretical analysis can be found in \cite{absil2009optimization,boumal2020intromanifolds}.

%Riemannian optimization techniques produce formal iterative algorithms concerning the geometry data. In these algorithms, the optimization method works according to the geodesic curve of the manifold that is related to the geometry of the problem. In the case of working with low-rank manifolds, the geometry of SVD is used \cite{vandereycken2013low}. A complete discussion of Riemannian optimization methods with theoretical analysis can be found in \cite{absil2009optimization,boumal2020intromanifolds}.

One of the instances of low-rank learning that Riemannian optimization can be used to solve is Riemannian similarity learning (RSL) that we explain here. Suppose we have pairs of data from following two domains
%there are two different (or similar) sources of data with the following domains
\begin{subequations}
\label{eq:Sim_Dom}
\begin{align}
     \label{eq:Sim_Dom1}
        \mathcal{D}_{X}&=\{\boldsymbol{x_1},\boldsymbol{x_2},...\}, \boldsymbol{x_i}\in \mathbb{R}^{d_1},\\
        \mathcal{D}_{V}&=\{\boldsymbol{v_1},\boldsymbol{v_2},...\},\boldsymbol{v_j}\in \mathbb{R}^{d_2},
\end{align}
\end{subequations}
and the task is estimating a function
%The predictive model is defined as a learning problem with the form of
$\hat{y}:\mathcal{D}_{X}\times\mathcal{D}_{V}\to{\mathbb{R}}$ that measures the similarity of these pair of data, where '$\times$' represents the Cartesian product of two sets. The training data consists of $n$ tuples $(\boldsymbol{x_i},\boldsymbol{v_i},y_i),\quad i\in \{1,2,...,n\}$, where $\boldsymbol{x_i}\in \mathcal{D}_{X}$, $\boldsymbol{v_i}\in \mathcal{D}_{V}$ and $y_i$ is a value measuring the similarity of the data pair $(\boldsymbol{x_i},\boldsymbol{v_i})$.
In \cite{chechik2009online,chechik2010large}, the following parametric function was introduced for measuring the similarity:
%\par From a regression point of view, $ y_i\in \mathbb{R}$ and for new upcoming test pair $(\boldsymbol x,\boldsymbol v,y)$, the predictive model can be defined as the following
%$\hat{y}=f_W$ such that $f_W$
\begin{equation}
\label{eq:f_w}
\begin{aligned}
f_W : (\boldsymbol{x},\boldsymbol{v})\mapsto\boldsymbol{x}^T\boldsymbol{W}\boldsymbol{v}.
 \end{aligned}
\end{equation}
%where $f_W$ is an estimation of $y$.
%The function $f_W$ was proposed in  \cite{chechik2009online,chechik2010large} for the first time.
%In \cite{chechik2009online,chechik2010large}, $\boldsymbol x,\boldsymbol v \in \mathbb{R}^d$ were assumed to be sparse vectors, that is
%\begin{equation}\label{x_d_sparse}
%\|\boldsymbol x\|_0=\|\boldsymbol v\|_0\ll d,
%\end{equation}
%where $\|.\|_0$ defined as the number of nonzero elements.
In \cite{mishra2014r3mc}, similar cost as that of \eqref{eq:f_w} was used for similarity learning but they induces low-rank constraint on $W$ to regularize their model.
%After that, a new form of \eqref{eq:f_w} that was related to the numerical rank of $W$ was introduced in \cite{mishra2014r3mc}. In this new form, $W$ was a low-rank matrix.
In some problems, $ y_i\in \{-1,1\}$ depending on whether the two samples are similar or not. Therefore, similarity learning can be cast as the following problem:
\begin{equation}\label{bi_prob}
   \min_{W\in \mathcal{M}_r}
    f(W) =1/n\sum_{i=1}^nl(f_W(\boldsymbol{x}_i,\boldsymbol{v}_i),y_i),
\end{equation}
where $l$ is a loss function such as hinge or cross-entropy loss function, and $\mathcal{M}_r=\{W\in \mathbb{R}^{d_1\times d_2} : rank(W)=r\}$. Since the low-rank constraint can be seen as a Riemannian manifold, this similarity learning problem can be called Riemannian similarity learning.
%With the above assumptions the following Riemannian optimization problem can be proposed for the problem of similarity learning between data pairs of two separate datasets:
To solve \eqref{bi_prob}, one of the different types of Riemannian gradient descent methods presented in \cite{bonnabel2013stochastic,kasai2018riemannian,zhang2018robust} can be used. We assume that the rank is significantly smaller than the matrix dimensions, i.e. $r\ll \textbf{min}(d_1,d_2)$. In addition, to improve the performance, we use Algorithm \ref{alg_fast_SVD} inside a Riemannian gradient descent algorithm instead of standard SVD to solve \eqref{bi_prob}.

%{\color{red}
%In this section, we explain an application of our method for low-rank learning.
%
%One of the common approaches for low-rank learning is using Riemannian stochastic gradient descent (RSGD)\cite{bonnabel2013stochastic}.
%
%We use manifold of low-rank matrices in our optimization process \cite{vandereycken2013low} which is based on SVD. To accelerate this process, we perform the proposed Algorithm \ref{alg_fast_SVD} instead of standard SVD, and evaluate the results.
%}

\subsection{Riemannian stochastic gradient descent}
Consider the following minimization problem:
\begin{equation}\label{rsgd_opt_func}
   \min_{W\in \mathcal{M}_r}
    f(W),
\end{equation}
where $\mathcal{M}_r$ is the Riemannian manifold of matrices with rank equal to $r$, and we use a noisy version of the Riemannian gradient ($\nabla f_{t}$) at step $t$ in the update rule. Riemannian stochastic gradient descent (RSGD) \cite{bonnabel2013stochastic} uses the following update rule:
\begin{equation}\label{rsgd_update}
   W_{t+1}=\text{R}_{W_t}(-\eta_t\nabla f_{t}),
\end{equation}
where $\eta_t$ is the step size at step $t$, and $\text{R}_{W_t}$ is a retraction at $W_t$. A retraction is a mapping from tangent space at a point to the manifold satisfying certain properties. Indeed for any tangent vector $\xi \in T_W\mathcal{M}_r$, a retraction can be found by solving the following equation \cite{vandereycken2013low}

\newcommand{\argmin}{\operatornamewithlimits{argmin}}

\begin{equation}\label{retraction_general}
   R_W(\xi)=\argmin _{X\in \mathcal{M}_r} \|W+\xi-X\|_F.
\end{equation}
This equation has a closed form solution as
\begin{equation}\label{retraction_svd}
   R_W(\xi)=\sum_{i=1}^r\sigma_iu_iv_i^T,
\end{equation}
where $\{\sigma_i,u_i,v_i\}_{i=1}^r$ are the first $r$ singular triplets of SVD in the point $W+\xi$.

\paragraph{Geometric meaning of SVD}
Suppose $\mathcal{M}_W:=\{W\in \mathbb{R}^{d_1\times d_2} :W=U\Sigma V^T\}$, where $U$, $\Sigma$, $V$ are matrices defined the same as \eqref{eq_svd}. It was demonstrated in \cite[Chapter~3]{absil2009optimization} and \cite{vandereycken2013low} that $\mathcal{M}_W$ is a Riemanninan manifold, and the tangent space at point $W$ is defined by
\begin{equation}\label{svd_tangent_vector}
   T_W\mathcal{M}:=\{UMV^T+U_PV^T+UV_P^T\},
\end{equation}
where $U_P^TU=0$ and $V_P^TV=0$, and $M$ is an arbitrary $r$ by $r$ matrix. The Riemannian metric is defined as $<\eta,\zeta>=\text{tr}(\eta^T\zeta)$, where $\text{tr}(.)$ is the matrix trace.

\paragraph{Riemannian gradient}
Assume $\text{Gr}_W^{f}\in \mathbb{R}^{m\times n}$ represents the Euclidean gradient of the objective function \eqref{rsgd_opt_func} w.r.t to $W$. The Riemannian gradient, that is computed by the projection of the Euclidean gradient onto the tangent space of $\mathcal{M}_W$, has the following form
\begin{equation}\label{svd_Riemannian_gradient}
\text{Grad}_Wf=P_U^\mathcal{H}\text{Gr}_W^{f}P_V^\mathcal{H}+P_U^\mathcal{V}\text{Gr}_W^{f}P_V^\mathcal{H}+P_U^\mathcal{H}\text{Gr}_W^{f}P_V^\mathcal{V},
\end{equation}
where the shorthand notations $P_X^\mathcal{H}:=XX^T$ and $P_X^\mathcal{V}:=I-XX^T$.

\paragraph{RSGD algorithm for RSL}
The complete process of the RSGD method that is customized for the Riemannian similarity learning problem is represented in Algorithm~\ref{alg_fast_rsgd}.% In this algorithm., $Gr$ represents Euclidean gradient.
\\

\begin{algorithm}
\caption{Fast Riemannian mini-batch gradient descent algorithm} \label{alg_fast_rsgd}

\textbf{Inputs:}
\\\hspace*{10pt}$X \in \mathbb{R}^{m \times d_1}$, $V \in \mathbb{R}^{n \times d_2}$,
\\\hspace*{10pt}$K$ : number of iterations, b: mini batch size
\\\hspace*{10pt} $\eta$: RSGD training rate
\\\hspace*{10pt} $r$: rank of low-rank manifold $\mathcal{M}$
\\\textbf{Outputs:}
\\\hspace*{10pt}$W\in \mathbb{R}^{d_1 \times d_2}$ as the optimized parameter matrix,
\\\hspace*{10pt}

\begin{algorithmic}[1]
\State $W\sim\mathcal{N}(0,1)^{d_1\times d_2}$
%\State t=0
\State  For each step $t<K$ do
\State \hspace*{10pt}$Gr \in \mathbb{R}^{d_1 \times d_2}=0$
\State \hspace*{10pt} Draw a batch of data tuple $\mathcal{B}=\{(\boldsymbol{x}_i,\boldsymbol{v}_i,y_i)\}_{i=1}^b$
\State \hspace*{10pt} Compute
 \hspace*{30pt}$Gr=\frac{1}{|\mathcal{B}|}\sum\limits_{(\boldsymbol{x}_i,\boldsymbol{v}_i,y_i)\in \mathcal{B}}(\nabla f_W(\boldsymbol{x}_i,\boldsymbol{v}_i))y_i$
\State \hspace*{10pt}$Gr = Gr - \lambda*W$
\State \hspace*{10pt}Compute the SVD of $Gr=U_r \Sigma_r V^T_r$ using Algorithm~\ref{alg_fast_SVD} with desired rank $r$
\State \hspace*{10pt}$Z=U_rU_r^TGrV_rV_r^T+(I-U_rU_r^T)GrV_rV_r^T+U_rU_r^TGr(I-V_rV_r^T)$
\State \hspace*{10pt}Compute the SVD of ($W-\eta Z)=U_r \Sigma_r V_r$ using Algorithm~\ref{alg_fast_SVD} with desired rank $r$
\State \hspace*{10pt}$W_{new}=U_r \Sigma_r V_r$
\State end For
\end{algorithmic}
\end{algorithm}

\section{Experimental results}
%The experimental results are presented in three parts.
\label{section_experiments}
%\subsection{Experimental setting}
%\label{section_exp_setting}

In this section we investigate the effectiveness of Algorithms \ref{alg_fast_rank}, \ref{alg_fast_SVD}, and \ref{alg_rank_determination} which we developed in Python 3.7.6. Our experiments were performed on the google cloud with 16 vCPUs @ 2.2 GHz (8 real cores) and 128GB RAM. To measure execution times and errors, we calculate an average value for the results of five repetitions of each algorithm. Our Python code is available via \url{https://github.com/rezagodaz/accurate-partial-svd} .

\begin{table*}
\caption{Comparison of various SVD algorithms with respect to execution time (a), estimated numerical rank of the input matrix (b), and the observed residual and relative errors (c). The results are obtained using synthetic matrices of different sizes, all of which have a numerical rank equal to 100. For the randomized SVD algorithm (R-SVD) of \cite{halko2011finding}, we investigate two possible scenarios: (i) knowing the required value of oversampling parameter $p$ (see section \ref{section_randomized}), and (ii) using the default value of $p$ by \cite{halko2011finding}). For (b) and (c), the underlying goal is to determine the 20 dominant triplets of the input matrix using the various possible SVD algorithms and to compare them with respect to execution time and accuracy.  }
\label{table_random}
\begin{subtable}{.5\linewidth}
\caption{}
\label{table_times_a}
\centering
\setlength{\tabcolsep}{+2pt}
\begin{tabular}{|c|c|c|c|c|}
\hline
\begin{tabular}[c]{@{}c@{}}Input \\ Matrix\\ size\end{tabular} &
  \multicolumn{3}{c|}{\begin{tabular}[c]{@{}c@{}}Rank determination \\ Time\\ (sec)\end{tabular}} &
  \begin{tabular}[c]{@{}c@{}}Alg \ref{alg_fast_rank} \\ number of\\ iterations\end{tabular} \\ \cline{2-4}
        & SVD     & Alg \ref{alg_fast_rank}  & Alg \ref{alg_rank_determination}  &     \\ \hline
$1e3\times1e3$ & 0.17    & 0.061   & 0.063   & 102 \\ \hline
$1e4\times1e3$ & 0.59    & 0.639   & 0.641   & 102 \\ \hline
$1e5\times1e3$ & 5.92    & 6.588   & 6.590   & 102 \\ \hline
$1e4\times1e4$ & 113.52  & 4.070   & 4.073   & 104 \\ \hline
$1e5\times1e4$ & 379.62  & 32.548  & 32.550  & 104 \\ \hline
$1e5\times2e4$ & 1854.72 & 59.134  & 59.137  & 104 \\ \hline
$1e5\times3e4$ & 4788.48 & 77.530  & 77.533  & 105 \\ \hline
$1e5\times8e4$ & NA      & 185.541 & 185.544 & 105 \\ \hline
\end{tabular}
%\caption{Time evaluation of the\\ rank finder algorithms}
\end{subtable}
\begin{subtable}{.5\linewidth}
\centering
\setlength{\tabcolsep}{+2pt}
\caption{}
\label{table_times_b}
\begin{tabular}{|c|c|c|c|c|}
\hline
\multirow {2}{*}{\begin{tabular}[c]{@{}c@{}}Input \\ Matrix\\ size\end{tabular}} &
  \multicolumn{4}{c|}{\begin{tabular}[c]{@{}c@{}}Time of Algorithms\\(sec)\end{tabular}} \\ \cline{2-5}
 &
  SVD &
  F-SVD &
  \begin{tabular}[c]{@{}c@{}}R-SVD\\ (default)\end{tabular} &
  \begin{tabular}[c]{@{}c@{}}R-SVD\\ (oversampled)\end{tabular} \\ \hline
$1e3\times1e3$ & 0.33    & 0.06   & 0.03   & 0.10   \\ \hline
$1e4\times1e3$ & 1.16    & 0.53   & 0.24   & 0.65   \\ \hline
$1e5\times1e3$ & 10.27   & 5.29   & 3.15   & 11.02  \\ \hline
$1e4\times1e4$ & 163.06  & 3.18   & 2.02   & 5.64   \\ \hline
$1e5\times1e4$ & 856.55  & 28.60  & 23.83  & 52.15  \\ \hline
$1e5\times2e4$ & 4520.32 & 65.84  & 63.44  & 102.18 \\ \hline
\multicolumn{1}{|l|}{$1e5\times3e4$} &
  NA &
  \multicolumn{1}{l|}{77.73} &
  69.83 &
  142.45 \\ \hline
$1e5\times8e4$ & NA      & 203.18 & 201.97 & 362.04 \\ \hline
\end{tabular}
%%\caption{Time evaluation of the singular triplet finder algorithms.\\ (na=not applicable in the execution setting)}
\end{subtable}%
\vspace{0.3cm}
\\
\begin{subtable}{1\linewidth}
\centering
\caption{}
\label{table_relative_err}
\begin{tabular}{|c|l|l|c|l|c|l|c|l|}
\hline
\begin{tabular}[c]{@{}c@{}}Input \\ Matrix\\ size\end{tabular} &
  \multicolumn{2}{c|}{\begin{tabular}[c]{@{}c@{}}error of\\ SVD\end{tabular}} &
  \multicolumn{2}{c|}{\begin{tabular}[c]{@{}c@{}}error of \\ F-SVD\end{tabular}} &
  \multicolumn{2}{c|}{\begin{tabular}[c]{@{}c@{}}error of\\ R-SVD\\ (oversampled)\end{tabular}} &
  \multicolumn{2}{c|}{\begin{tabular}[c]{@{}c@{}}error of\\ R-SVD\\ (default)\end{tabular}} \\ \cline{2-9}
 &
  \multicolumn{1}{c|}{Residual} & Relative & Residual & Relative & Residual & Relative &
  \multicolumn{1}{l|}{Residual} & Relative \\ \hline
$1e3\times1e3$ & \textbf{6.97e-12} & 1.90e-15 & 6.77e-11 & \textbf{7.27e-17} & 2656.72 & 2.28e-15 & 2664.66 & 1.35e-15 \\ \hline
$1e4\times1e3$ & \textbf{9.89e-12} & 2.83e-15 & 8.01e-11 & \textbf{7.43e-17} & 2660.99 &  2.69e-15 & 2664.57 & 1.56e-15 \\ \hline
$1e5\times1e3$ & \textbf{9.97e-12} & 2.96e-15 & 2.24e-10 & \textbf{7.26e-17} & 2646.96 & 2.36e-15 & 2665.86 & 1.68e-15 \\ \hline
$1e4\times1e4$ & \textbf{3.20e-11} & 3.14e-15 & 3.92e-10 & \textbf{8.04e-17} & 8783.91 & 1.86e-15 & 8810.52 & 2.00e-15 \\ \hline
$1e5\times1e4$ & \textbf{4.51e-11} & 4.25e-15 & 3.73e-10 & \textbf{8.56e-17} & 8779.20 & 2.03e-15 & 8616.37 & 1.72e-15 \\ \hline
$1e5\times2e4$ & \textbf{6.69e-11} & 4.50e-15 & 3.07e-10 & \textbf{7.06e-17} & 12494.80 & 2.03e-15 & 12543.44 & 1.80e-15 \\ \hline
$1e5\times3e4$ & \multicolumn{1}{c|}{NA} & \multicolumn{1}{c|}{NA} & \multicolumn{1}{l|}{2.30e-9} & \textbf{8.18e-17} & 15332.69 & 2.06e-15 & 15399.78 & 1.87e-15 \\ \hline
$1e5\times8e4$ & \multicolumn{1}{c|}{NA} & \multicolumn{1}{c|}{NA} & NA & \textbf{7.30e-17} & NA & 2.03e-15 & NA & 1.74e-15 \\ \hline
\end{tabular}
\end{subtable}
\end{table*}

\begin{figure}[t]
    \centering
    \includegraphics[width=8.8cm]{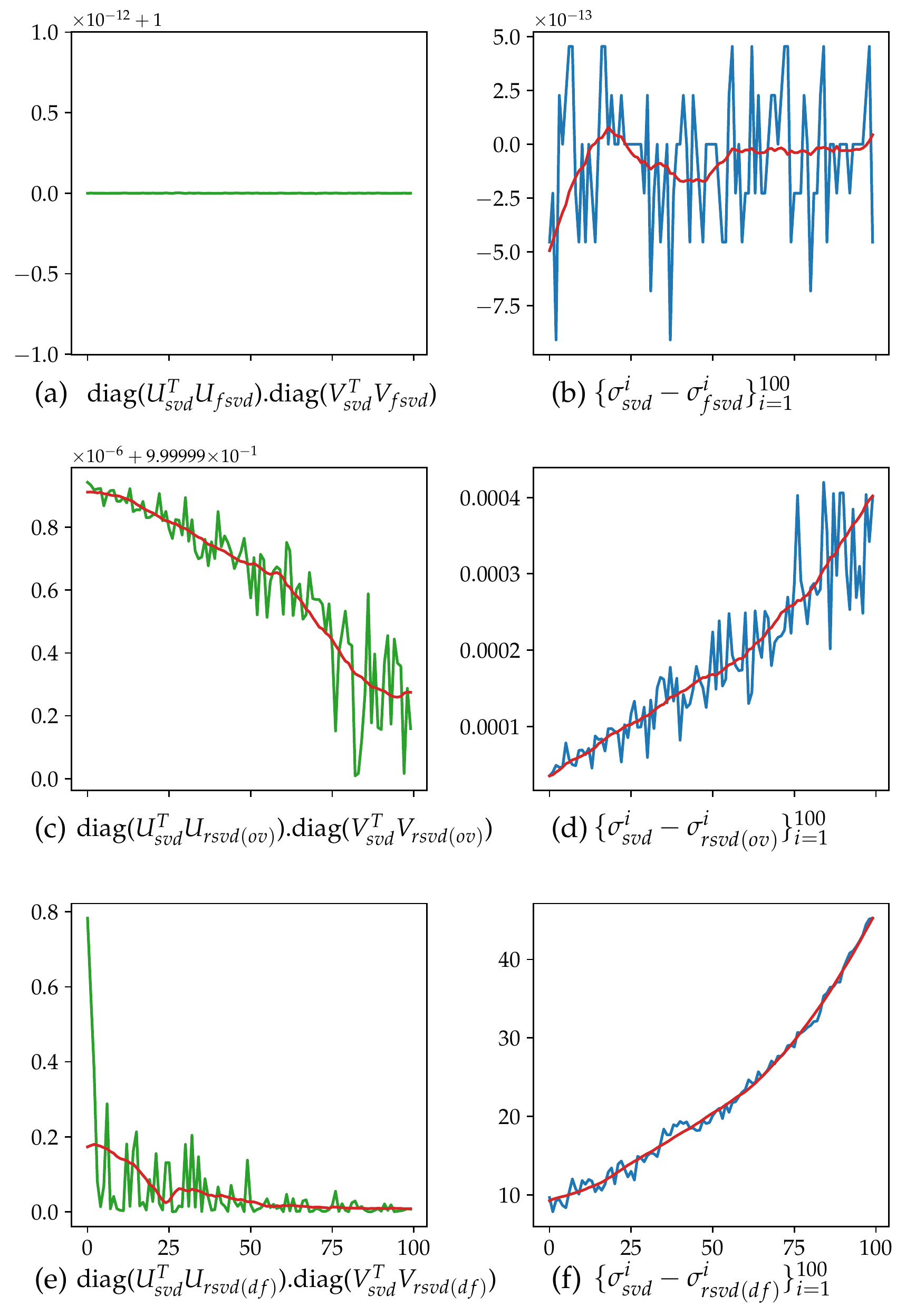}
    \caption{ Investigating the quality of the singular triplets of our proposed F-SVD (c and d), oversampled R-SVD (e and f) and default R-SVD (e and f)  compared to the singular triplets found by traditional SVD. The input matrix is $A\in \mathbb{R}^{1e4 \times 1e4}$ with the numerical rank equal to 1000 and the goal is to determine 100 dominant singular triplets, and their singular values by an SVD algorithm:  $A\approx U_{alg}{\Sigma_{alg}}V_{alg}^T$. In the plots of the first row, we are showing the values of the error vector computed by $diag(U^T_{svd}U_{alg}).diag(V^T_{svd}V_{alg})$, where $alg$ denotes one of the three SVD methods,  $svd$  is the standard SVD method used by the Numpy Python library, $diag(.)$ returns diagonal elements of a matrix as a vector, and dot represents element-wise multiplication of two vectors. In the plots of the second row, the values of the absolute difference between dominant singular values $\{ |\sigma_{svd}^i-\sigma_{alg}^i|\}_{i=1}^{100}$ are shown. The red curves in the plots is the smoothed version of the original curves.
      }

    \label{fig_comparison}
\end{figure}

\subsection{Comparing accuracy and speed of SVD algorithms}
\label{section_exp_results}

We select two algorithms, traditional SVD and R-SVD \cite{halko2011finding} for our comparisons because traditional SVD is one of the most accurate methods and R-SVD is one of the fastest choices among the randomized methods. In addition the performance of R-SVD with respect to errors makes it a suitable representative for the various randomized methods.

The experimental results are shown in Table \ref{table_random}. In this experiment to build a synthetic matrix $A\in \mathbb{R}^{m\times n}$, with fixed rank $l$, we multiplied two matrices $M\in \mathbb{R}^{m\times l}$ and $N\in \mathbb{R}^{l\times n}$. Each instance of $M$ and $N$ was randomly and independently using a Gaussian distribution. The values of $m$ and $n$ vary as included in the first columns of Tables \ref{table_times_a}, \ref{table_times_b}, and \ref{table_relative_err}. To determine the error of each algorithm we define two kinds of errors. The errors reported in Table \ref{table_relative_err} are defined as the relative error, $err_{rel}=\|A^TU-V\Sigma\|_{F}/\|\Sigma\|_{F}$, and the residual error,  $err_{res}=\|A-U\Sigma V^T\|_{F}$, where the matrices $U$, $\Sigma$, and $V$ are generated using the specified algorithms given in the tables.

Table \ref{table_times_a} is related to approximate the numerical rank of a huge matrix that is a secondary result of Algorithm \ref{alg_fast_rank}. The proposed Algorithm \ref{alg_rank_determination} is a fast algorithm to determine the numerical rank of the input matrix that providing a distinct time advantage over the current practical method used by Python, which executes traditional partial SVD, for the situations that input matrix is huge. When the true numerical rank of a huge original matrix is small, Algorithm \ref{alg_fast_rank} and also Algorithm \ref{alg_rank_determination} perform the best. The last column of Table \ref{table_times_a} records the number of iterations of Algorithm \ref{alg_rank_determination} after terminating based on the condition $\|q_{k'}\|<\epsilon$ in iteration $k'$, without any predefined parameter or user intervention. This value is our first approximation of the numerical rank of the input matrix (In Table \ref{table_times_a}, Alg \ref{alg_fast_rank} and Alg \ref{alg_rank_determination} denote the algorithms \ref{alg_fast_rank} and \ref{alg_rank_determination}, respectively).

%
%\begin{table*}[!t]
%\centering
%\caption{Comparing the observed residual and relative errors of four SVD algorithms.}
%
%\label{table_relative_err}
%\end{table*}

%\clearpage

Considering the complexity analysis of the F-SVD method (mentioned in Section \ref{section_complexity_F-SVD}) and also the experimental results summarized in Table \ref{table_times_b},  F-SVD  has a computational complexity comparable to that of R-SVD when the decay of matrix singular values is slow. In addition, it can be seen that the execution time of the F-SVD methods is less than that of R-SVD (oversampled) method.

Table  \ref{table_relative_err}  shows that Algorithm \ref{alg_fast_SVD} exhibits  the  smallest relative error among all executions. It also gives low residual error and fast computation of the singular values and the corresponding singular vectors, especially the smaller singular triplets, which represents the relative superiority of F-SVD over R-SVD (oversampled).Therefore, this algorithm is a suitable choice for use in almost all problem settings.

%(see Table \ref{table_relative_err} and also Figure \ref{fig_comparison}).

In Figure \ref{fig_comparison},  we are comparing the quality of SVD triplet obtained by different methods. In this experiment, Algorithm \ref{alg_fast_SVD} terminates after 550 iterations. For the algorithm R-SVD, the oversampling parameter was set to 800 (p=800 thus l=900). For the plots in the first row of this figure, a value close to 1.0 means that the corresponding vectors computed were satisfactorily accurate. It can be seen that Algorithm \ref{alg_fast_SVD} outperforms its R-SVD competitor with respect to the accuracy for both the singular values as well as the singular vectors.

\subsection{RSL application}
\begin{figure}
\centering{\includegraphics[width=8.8cm]{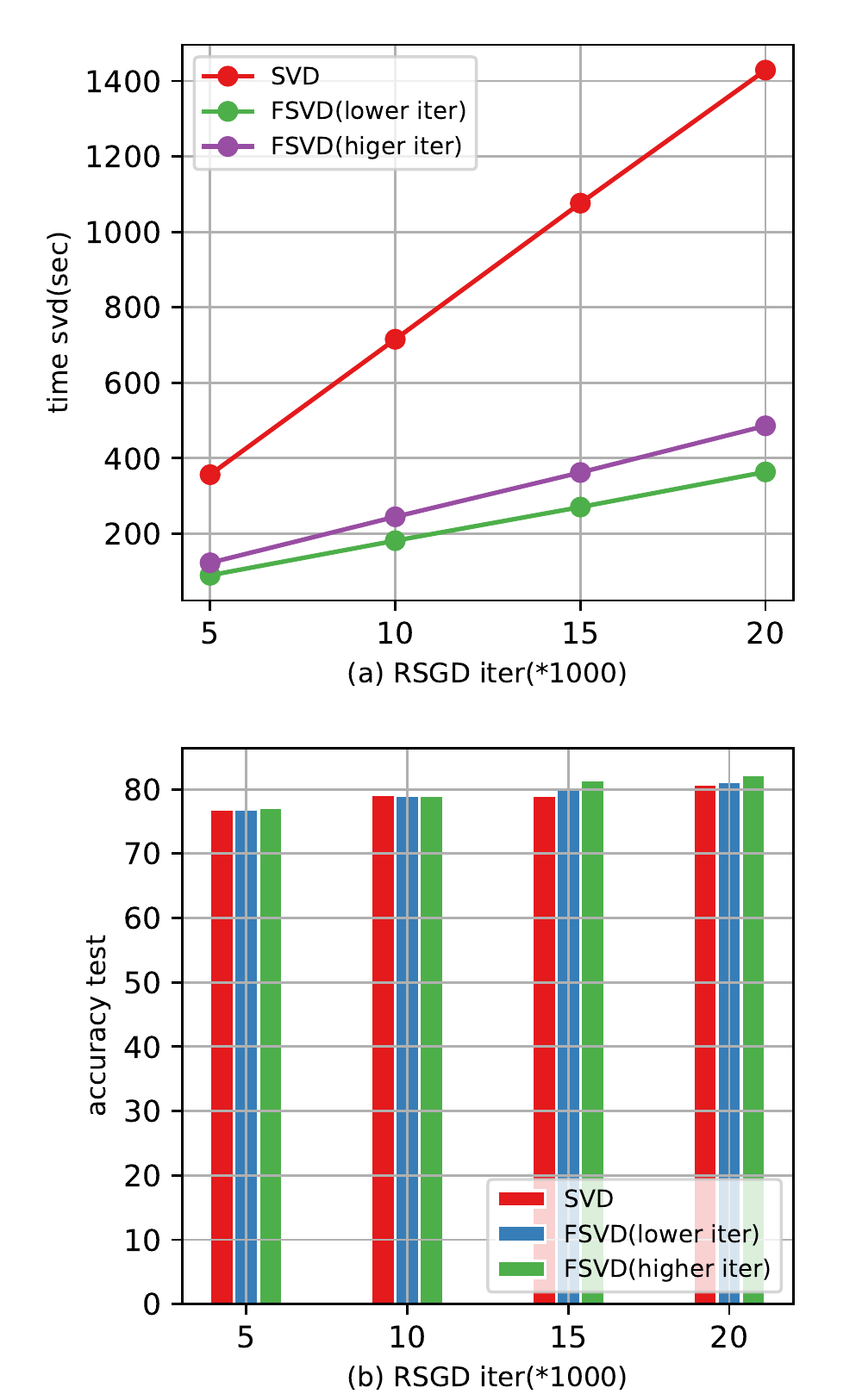}}
\caption{The execution time (a) and the test accuracy (b) of the RSL algorithm between two handwritten digit  datasets (MNIST and USPS) for different number of iterations of the RSGD algorithm. Algorithm~\ref{alg_fast_rsgd} is used for RSL with three different SVD methods. These three SVD methods are: (i) standard SVD method, (ii) our proposed F-SVD method with setting inner iteration of  Algorithm~\ref{alg_fast_SVD} to 20 ("lower iter"), and (iii) F-SVD method with setting inner iteration to 35 ("higher iter").
}\label{fig_comparison_rsgd}
\end{figure}

In this section, we evaluate Algorithm~\ref{alg_fast_rsgd} in the task of similarity learning for classification. We use two famous datasets MNIST (used as the source domain $\mathcal{D}_X$) and USPS (used as the target domain $\mathcal{D}_V$). Both of these datasets include handwritten digits, but with a different number of pixels. Algorithm~\ref{alg_fast_rsgd} is used for the task of RSL with the input parameter rank equal to five, and the number of iterations varying from 5000 to 20000. Figure \ref{fig_comparison_rsgd} compares the running-time and accuracy results obtained by using the standard SVD and FSVD in the inner loop of Algorithm~\ref{alg_fast_rsgd}. As it is clear, using FSVD significantly leads to less computation time than the standard SVD, while the same accuracy is obtained.
The results in this figure is the median result for three executions of Algorithm~\ref{alg_fast_rsgd}. It should be mentioned that Riemannian optimization algorithms such as that of Algorithm \ref{alg_fast_rsgd} need a high level of accuracy for the retraction and the RSVD method can not be used in these types of algorithms.

%\\

%\clearpage

\section{Conclusion}
\label{section_conclusion}
Traditional method for solving SVD are accurate for matrix decomposition that has many applications in different sciences. This algorithm is not a suitable choice for huge matrices because of its computational complexity. Randomized algorithms are a group of methods that tackle this problem by decomposing a smaller matrix made from the main matrix. They successfully reduced the computational complexity of the traditional SVD algorithm and hence can be used with huge input matrices. However, such randomized algorithms may be inaccurate, or may require unknown values for their key parameters. Using key concepts from Krylov subspaces, we have devised SVD algorithms that execute quickly and provide accurate, reliable singular values and their corresponding singular vectors for huge input matrices. As a by-product, we obtained a fast and accurate rank estimation for huge matrices.

\section*{Acknowledgment}
The authors thank Prof David Matthews, a faculty of statistics and actual science at the University of Waterloo, Canada for carefully proofreading the manuscript.
%\begin{IEEEbiographynophoto}{}


%\vfill

\bibliographystyle{IEEEtran}
\bibliography{references}

\end{document}